# Degree of Irrationality: Sentiment and Implied Volatility Surface

Jiahao WENG[a], Yan XIE[b]

[a]Physics Department, Renmin University of China, Beijing, 100872, Beijing, China
[b]School of Finance, Renmin University of China, Beijing, 100872, Beijing, China

**Abstract**

In this study, we constructed daily high-frequency sentiment data and used the VAR method to attempt to predict the next day's implied volatility surface. We utilized 630,000 text data entries from East Money Stock Forum from 2014 to 2023, and employed deep learning methods such as BERT and LSTM to build daily market sentiment indicators. By applying FFT and EMD methods for sentiment decomposition, we found that high-frequency senti- ment had a stronger correlation with at-the-money (ATM) options' implied volatility, while low-frequency sentiment was more strongly correlated with deep out-of-the-money (DOTM) options' implied volatility. Further analysis revealed that the shape of the implied volatility surface contains richer market sentiment information beyond just market panic. We demon- strated that incorporating this sentiment information can improve the accuracy of implied volatility surface predictions.

*Keywords:* implied volatility surface, market sentiment, deep learning

## 1. Introduction

Current research generally considers the options market to contain a large number of informed traders with insider knowledge and more knowledgeable traders due to the high leverage characteristics of the options market Easley et al. (1998). As such, indicators in the options market, such as options prices, implied volatility, and the Greeks, are seen as "smarter" compared to indicators in the securities market. Numerous studies have confirmed this perspective and have explored the discovery function of options implied volatility on se- curities prices. For instance, Ni et al. (2020) found that the degree of skewness in implied volatility smiles has a significant predictive ability for stock market returns, while Han and Li (2021) discovered that the difference between call and put implied volatility has signifi- cant predictive power for stock market returns. Additionally, there is more research on the predictive ability of options implied volatility on realized volatility, dating back to Latane and Rendleman (1976-05) reverse use of the BS formula to derive the implied standard de- viation of options and constructing a weighted implied standard deviation (WISD) using delta-neutral weighting, which was found to predict actual volatility significantly better than methods based on historical volatility. In recent years, numerous studies have incorporated the VIX index and the HAR method proposed by Corsi (2009), achieving notable results in predicting stock market volatility Byun and Kim (2013); Zhang (2020); Wan and Tian (2023).

However, indicators in the options market should not be treated as the gold standard. In 1972, Black and Scholes (1972) proposed the famous options pricing formula, which outlined the factors determining options prices such as underlying price, risk-free rate, strike price, time to expiration, and volatility. However, they also emphasized that the effectiveness of the BS formula largely depends on estimating the aforementioned parameters, and there can be a significant amount of investor bias during the estimation period. When explaining the characteristics of options in 1975, Black also reiterated that estimating volatility is challenging when considering taxes, transaction costs, and various constraints. Mahani and Poteshman (2008) discovered the phenomenon of overreaction in the options market and found that the implied volatility smile is influenced by market sentiment. Han (2008) found that the degree of the implied volatility surface smile is related to sentiment, with greater smile curvature during periods of heightened market sentiment. Seo and Kim (2015) found that implied volatility has different volatility predictive abilities depending on the level of investor sentiment, due to the overreaction of the stock market during high sentiment periods causing misestimation of implied volatility. Market sentiment is typically considered a window into such irrational behavior as it aggregates market expectation biases. The aforementioned empirical studies indicate that options pricing is not entirely rational.

Therefore, two questions become very important: What type of information does options implied volatility provide for behavioral finance? How can volatility be estimated more accurately for options pricing?

These two questions are not contradictory but rather two sides of the same coin. This idea is very important. Each investor has biases in asset pricing, and the market does not completely eliminate such errors. The discrepancy between the equilibrium price offered by the market and an idealized rational price, which is the sum of investors' erroneous views, is what constitutes market sentiment Baker and Wurgler (2006). Finance provides various formulas to estimate the true value ideally, but following Keynes's (1936) concept of the beauty contest, where one guesses others' choices, we see irrationality itself as valuable information. We suggest that estimating biased values can lead to greater profitability. Suppose an asset should be priced at 100 based on current information in a neutral market sentiment environment. Due to a heightened bullish sentiment in the market, the equilibrium price is set at 105. Ordinarily, value reversion would suggest shorting the asset; however, if we estimate a further increase in sentiment, we could face serious losses (e.g., the 2021 GameStop event). Thus, given the sentiment information, this approach would not be advisable. We should either hedge the sentiment risk or leverage the sentiment factor to obtain its premium. Therefore, we should identify the irrational information provided by options implied volatility and predict and estimate its trend to more accurately estimate volatility.

In terms of the information provided by options implied volatility, Rubinstein (1985) discovered during non-parametric estimation that the market overestimated the prices of out-of-the-money call options with short-term expirations, marking the first observation of the smile shape in the options volatility surface. Bates (1991) explained that this includes information about left-tail risk in the underlying asset, reflecting investors' expectations about future market movements. Pan (2002) found that expectations of jump risk manifest differently in out-of-the-money options, particularly deep out-of-the-money and near-the- money options, while Yan (2011) demonstrated the correlation between the slope of out-of- the-money options nearing expiration and jump risk. Chen et al. (2023) found that option prices integrate public and private information about stock return volatility, particularly information related to the arrival of news events. This is a key reason why implied volatility predicts future volatility more effectively than other forecasting methods such as historical realized volatility, GARCH models, and stochastic volatility models. News channels account for approximately one-third of the overall predictive ability of implied volatility for future realized volatility.

In terms of estimating volatility more accurately, numerous studies have also found that the implied volatility surface not only presents a surface shape related to strike prices and expiration terms at a given time, but it also changes over time Cont and Da Fonseca (2002). Gonçalves and Guidolin (2006) discovered that the movement of the S&P 500 IVS is highly predictable, making the prediction of this change meaningful.

In recent years, machine learning has been involved in the prediction of IVS. Chen and Zhang (2019) introduced the attention mechanism into Long Short-Term Memory (LSTM) networks to establish a prediction system capable of capturing long-term memory of financial volatility, using the implied volatility smile surface of the S&P 500 index options market as the prediction target. The results indicated that the LSTM-Attention prediction system achieved error curve convergence and predicted the implied volatility surface more accurately than other prediction systems. Using a three-year rolling prediction of the implied volatility surface and applying the BS formula to price option contracts, they constructed time spread and butterfly spread strategies and found that the two strategies built using the predicted implied volatility surface yielded higher returns and Sharpe ratios. Medvedev and Wang (2022) applied Convolutional Long Short-Term Memory (ConvLSTM) neural networks to perform multi-step predictions of the S&P 500 implied volatility surface, finding that ConvLSTM outperformed LSTM and traditional time series models in terms of training data fit, with a significantly lower mean absolute percentage error than LSTM. Additionally, Medvedev and Wang (2022) also explored how incorporating historical spatial dynamic information of the IVS into the prediction model can significantly improve prediction accuracy.Dierckx et al. (2020) successfully predicted next-day changes in implied volatility using a random forest algorithm and found that attention and sentiment features extracted from Twitter could significantly enhance prediction performance.

However, the highly nonlinear structure of neurons in deep learning leads to deep learning appearing as a black box Samek et al. (2017), where we obtain better results but do not un- derstand the underlying reasons. Barredo Arrieta et al. (2020) pointed out that an important path to interpretable machine learning is simplifying the model and breaking down complex multi-step learning processes into multiple relatively simple parts.

In this paper, we constructed daily market sentiment indicators and achieved long-term and short-term sentiment decomposition using methods such as FFT and EMD. We found that high-frequency sentiment is more strongly correlated with at-the-money (ATM) options' implied volatility, whereas low-frequency sentiment is more strongly correlated with deep out- of-the-money (DOTM) options' implied volatility. This finding provides a new perspective on understanding the impact of market sentiment on option pricing. We improved the traditional VAR method by incorporating market sentiment indicators, which enhanced the accuracy of predicting the next day's implied volatility surface.

By leveraging natural language processing (NLP) technology, we innovatively combined market sentiment with implied volatility surface prediction. Utilizing text data from the East Money Stock Forum and processing it with deep learning methods such as BERT and LSTM, we significantly improved the accuracy of implied volatility surface predictions. Particularly, when using the BERT-BigBird model for sentiment analysis, the prediction performance surpassed other methods, demonstrating the effectiveness of deep learning techniques in financial forecasting. Compared to existing research, our project not only improves prediction accuracy but also emphasizes the logical interpretability of the model. By simplifying the model structure, we avoided the "black box" issue often associated with deep learning models, making the prediction results more transparent and understandable. Our empirical study confirmed the impact of market sentiment on the shape of the implied volatility surface, further corroborating Han's (2008) findings on the relationship between the implied volatility surface and market sentiment.

This paper will discuss the relationship between high- and low-frequency market sentiment and the implied volatility surface from a financial perspective in section 2. And in section 3, it will introduce machine learning to estimate market sentiment and predict the next day's implied volatility surface.

## 2. High- and low-frequency Sentiment and IVS

*2.1. Data*

*2.1.1. Implied Volatility Surface*

Due to the ease of data acquisition, this study uses the options dataset for the Shanghai Stock Exchange 50 ETF (stock code 510050.SH) issued by the Shanghai Futures Exchange as the underlying asset. These SSE 50 ETF options are among the most actively traded derivatives in China. They are European-style options and are physically settled with the ChinaAMC SSE 50 ETF. The expiration date of SSE 50 ETF options is the last Wednesday of the trading month, followed by four additional months in a quarterly cycle (March, June, September, and December). The strike prices are set at a range of fixed levels to accommodate different market conditions and investor demands, typically covering a range from 130.0% above to 60.0% below the underlying asset price.

Using Latane and Rendleman (1976-05) method, the implied volatility for the day is determined by inverting the BS formula:

$$C(S,t) = S_0 \cdot N(d_1) - K \cdot e^{-r(T-t)} \cdot N(d_2)$$
$$P(S,t) = K \cdot e^{-r(T-t)} \cdot N(-d_2) - S_0 \cdot N(-d_1)$$
$$d_1 = \frac{\ln(\frac{S_0}{K}) + (r + \frac{\sigma^2}{2})(T-t)}{\sigma\sqrt{T-t}} \quad , d_2 = d_1 - \sigma\sqrt{T-t}$$

Where N is the cumulative standard normal distribution function. Implied volatility can be calculated using numerical methods. We do not use implied volatility data obtained directly from the database for several reasons: Typically, for each option expiration date, which is the last Wednesday of each month, the liquidity of the options expiring that month drops sharply. To avoid biases related to market micro-structure, options with less than a week until expiration are excluded; similarly, deep out-of-the-money options with poor liquidity are often excluded from the database, and then the implied volatility surface is calculated. Since we are interested in the sentiment "bias" contained in the implied volatility surface, we cannot overlook this part of the data.

The options data we use is at daily frequency, ranging from February 9, 2015, to December 31, 2023. Based on the above data points, we can create a complete implied volatility surface using cubic spline interpolation. We select expiration periods of 1, 3, 6, and 12 months and six moneyness levels (130.0%, 110.0%, 102.5%, 97.5%, 90.0%, and 60.0%) as 24 data points for the nonparametric description of the implied volatility surface.

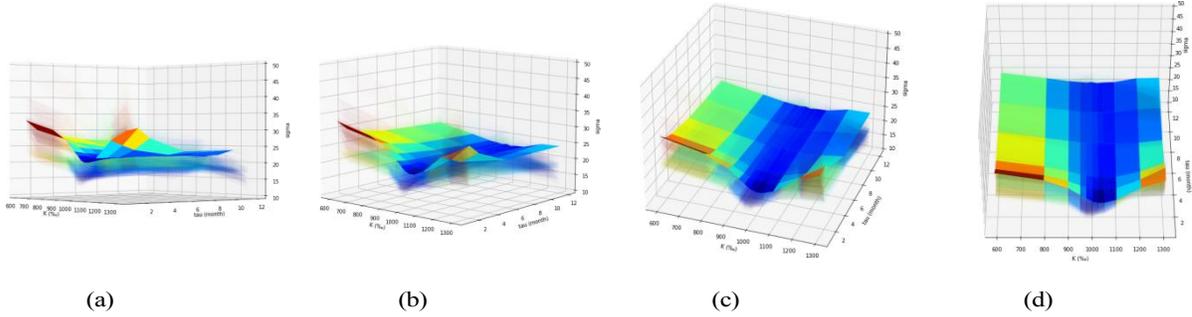

(a)        (b)        (c)        (d)

Figure 1: IVS

In addition, we can still use parametric methods to describe the implied volatility surface. In a cross-section of the same expiration period, especially for options with nearer expiration dates, the implied volatility can show a significant implied volatility smile. We can use curvature to characterize the degree of the smile:

$$\overline{Cur} = \frac{1}{n}\sum_{i=1}^{n-1} \frac{IV_{i-1} + IV_{i+1} - 2IV_i}{\Delta K^2 (1 + \frac{IV_{i+1} - IV_{i-1}}{2\Delta K})^{3/2}}$$

For left-right asymmetry in the implied volatility smile, we can examine its mean skewness:

$$\overline{Skew} = \frac{\frac{1}{n}\sum_i^n (IV_i - \overline{IV})^3}{\left[\frac{1}{n}\sum_i^n (IV_i - \overline{IV})^2\right]^{\frac{3}{2}}}$$

Descriptive statistics of the implied volatility surface are as follows:

*2.1.2. Sentiment*

The construction of sentiment indicators is the focus of this paper. Baker and Wurgler (2006, 2007) used principal component analysis to reduce the dimensionality of indicators such as the discount rate of closed-end funds, turnover rate, the number of IPOs, and the first-day return of IPOs, forming a sentiment indicator that has since been widely adopted and modified Zhang and Wang (2013). However, it has significant issues in terms of the difficulty of estimating at a daily frequency.

Currently, commonly adopted proxies for daily sentiment include the advance-decline line (ADL), turnover rate, and closed-end fund discount rate. The ADL reflects overall market trends by calculating the difference in the number of advancing and declining stocks each day to measure market sentiment:

$$ADL = N_{\text{up}} - N_{\text{down}} \tag{5}$$

Closed-end funds are investment companies that issue a fixed number of shares, which are then traded on stock exchanges. The discount (or occasionally premium) of closed-end funds is the difference between the net asset value of the fund's actual securities holdings and the market price of the fund. The turnover rate is a proxy for investor sentiment in the market, reflecting the frequency of stock trading; the higher the turnover rate, the higher the level of market sentiment. Zhang and Wang (2013) used the ratio of total market trading volume to circulating market capitalization to measure market turnover.

We use Baker and Wurgler's method to aggregate the above daily indicators into a daily sentiment index that we can obtain, which performs better than using an interpolation method with monthly data to obtain a daily index.

*2.2. Hypothesis and Model*

Han (2008) pointed out the correlation between the implied volatility smile surface and market sentiment, where the implied volatility smile surface describes the shape of implied volatility under different strike prices, and the "smile" shape reflects the market's view on dif- ferent risks. We aim to find the association between sentiment indices and the implied volatil- ity smile surface. Low-frequency sentiment is more correlated with at-the-money (ATM) op- tions, while high-frequency sentiment is more correlated with deep out-of-the-money (DOTM) options.

For a rational long-term investor, if they choose value investment, we tend to believe they would choose options with longer maturities to hedge their existing risks because shorter maturities are insufficient to hedge all risks. For a short-term investor, if they choose options with longer maturities, they cannot predict the returns when executing their short-term trading strategies because the prices of longer-term options are uncertain and still carry risks. To hedge risks and create new risks unnecessarily, we lean towards them choosing options with shorter maturities. Additionally, we believe that in assessing the magnitude of volatility, the weights of recent months should be higher, while the weights of distant months should be smaller. This might indicate that when the market sentiment seems to sharply rise in a particular month, when the high-frequency sentiment index is high, our estimation of its future might be higher than before, but not as high as the significant volatility of the most recent month. This association may reflect the impact of investor sentiment on market risk expectations.

Dumas et al. (1998) posits that volatility is not a fixed function of maturity ($\tau$), strike price (K), and underlying asset price (S), but rather a random variable.

Gonçalves and Guidolin (2006) established a vector autoregressive (VAR) model to cap- ture the time-series dynamics of implied volatility surface (IVS) coefficients. They first simulated parameters for a cross-section data using a quadratic function, then applied au- toregression to these parameters. They also compared the performance of the VAR model with other models such as the random walk model and the nonlinear GARCH model by Heston-Nandi, finding that VAR could better capture the temporal changes of IVS and achieve better predictive performance.

We expand the parameter vector of implied volatility for options, adding the current high and low-frequency sentiment indices for autoregression:

$$Y_t = \sum_k \Phi_k Y_{t-k} + \varepsilon_t$$

Here, when $Y_t = (Skew_{\tau_1}, Skew_{\tau_2}, \cdots, Cur_{\tau_1}, Cur_{\tau_2}, \cdots, Slope_{K_1}, Slope_{K_2}, \cdots, HFS, LFS)_t$ is applied, the equation takes a parameter form, whereas when $Y_t = (IV(\tau_1, K_1), IV(\tau_1, K_2), IV(\tau_1, K_3), \cdots, HFS, LFS)^T_t$ is applied, it takes a non-parameter form. For the parameter form, unlike the quadratic function modeling used by Gonçalves and Guidolin (2006) which can only represent smile degrees with quadratic coefficients, we directly describe the smile degree using Gaussian curvature, while also considering skew- ness at the same maturity, following Han's (2008) research. For the non-parameter form, we anticipate that low-frequency sentiment is more correlated with longer maturities and ATM options, while high-frequency sentiment is more correlated with shorter maturities and DOTM options.

*2.3. Method*

After obtaining daily sentiment, we decompose it into high-frequency and low-frequency components using various methods including Fourier decomposition, empirical mode decom- position (EMD), and moving average method.

*2.3.1. Fourier decomposition method*

The Fourier decomposition method is an important technique in signal analysis, which projects the signal onto the frequency-phase space:

$$\phi(\omega) = \frac{1}{\sqrt{2\pi}} \int IS(t) e^{-i\omega t} dt$$

In the phase space, we can significantly observe changes with frequency, with prominent behavior at low frequencies. We take this as a segmentation point, thus reconstructing the left side of the signal to form the low-frequency sentiment and reconstructing the right side of the signal to form the high-frequency sentiment.

$$\begin{cases} IS_{lfreq}(t) = \frac{1}{\sqrt{2\pi}} \int_0^{\arg\min_{\omega \in (0,N]} \phi(\omega)} \phi(\omega) e^{-i\omega t} dt \\ IS_{hfreq}(t) = \frac{1}{\sqrt{2\pi}} \int_{\arg\min_{\omega \in (0,N]} \phi(\omega)}^{\infty} \phi(\omega) e^{-i\omega t} dt \end{cases}$$

In the amplitude-frequency space, significant differences are observed near $\omega = 13$ and $\omega = 100$ in the sentiment curve, corresponding to frequencies of approximately 115 days and 15 days, respectively. These frequencies are typically associated with semiannual and monthly cycles, respectively, according to the usual economic cycle division. Considering the maturity of options on a monthly basis, in this study, we classify the portion of sentiment with $\omega < 100$ as high-frequency sentiment (HFS) and the portion with $\omega \leq 100$ as low-frequency sentiment (LFS), with $\omega \leq 13$ representing extreme low-frequency sentiment.

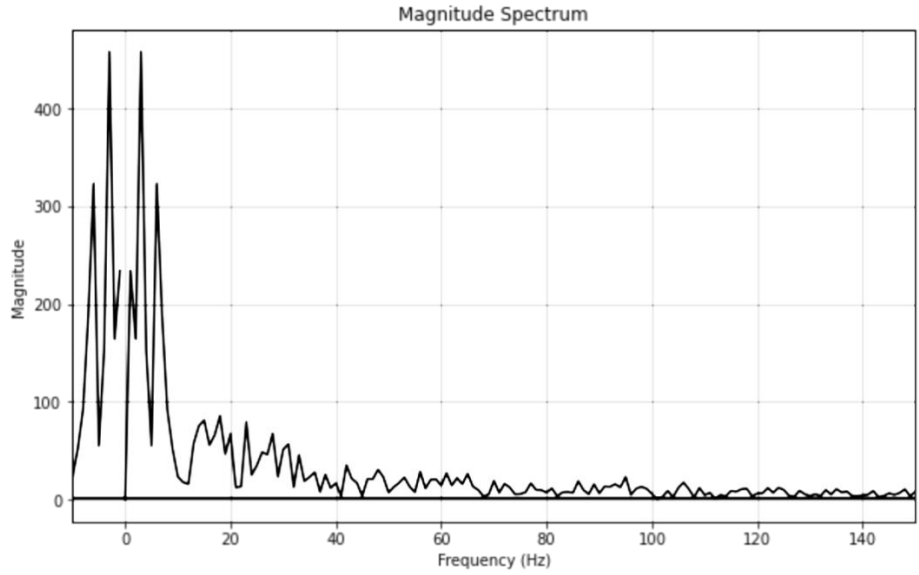

Figure 2: Panel 1: amplitude-frequency space

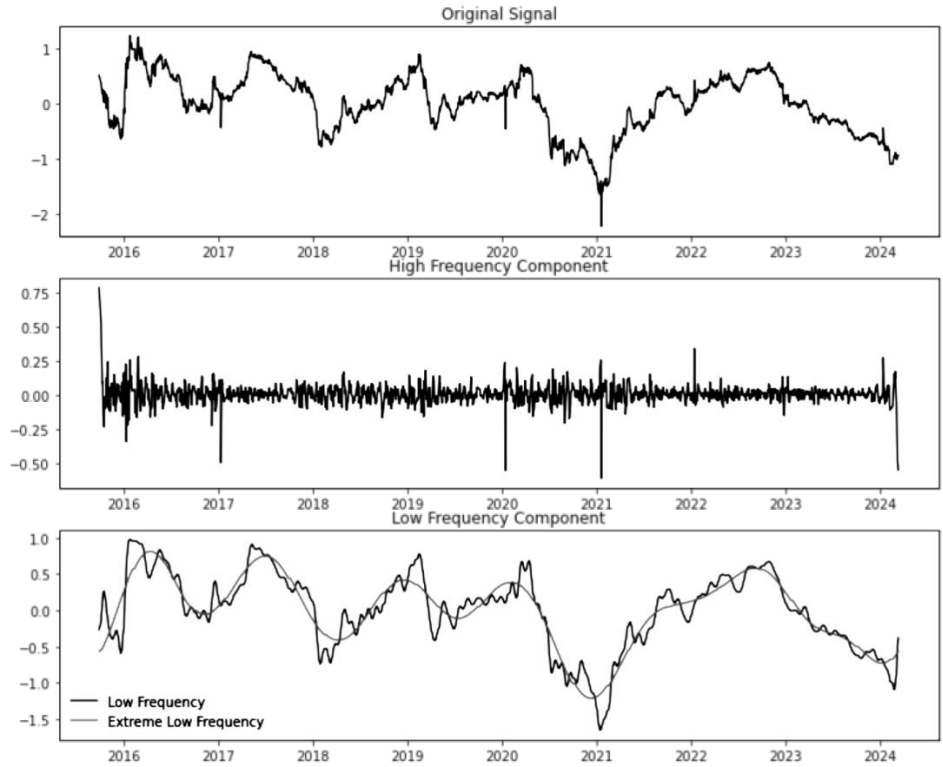

Figure 3: Panel 2: Real space

*2.3.2. Empirical Mode Decomposition*

Empirical Mode Decomposition (EMD) differs from Fourier analysis in that Fourier spec- tral analysis has inherent limitations in representing non-stationary data. Because Fourier spectra use uniform harmonic components for global analysis, many additional harmonic components are needed to simulate the non-stationarity or non-linear changes in the data. In contrast, EMD is an adaptive method that smooths the signal step by step by identifying extremum points Huang (2000). When the mean of the fifth intrinsic mode function (IMF) is significantly non-zero, we combine the first four IMFs to form high-frequency sentiment, while the remaining IMFs are combined to form low-frequency sentiment.

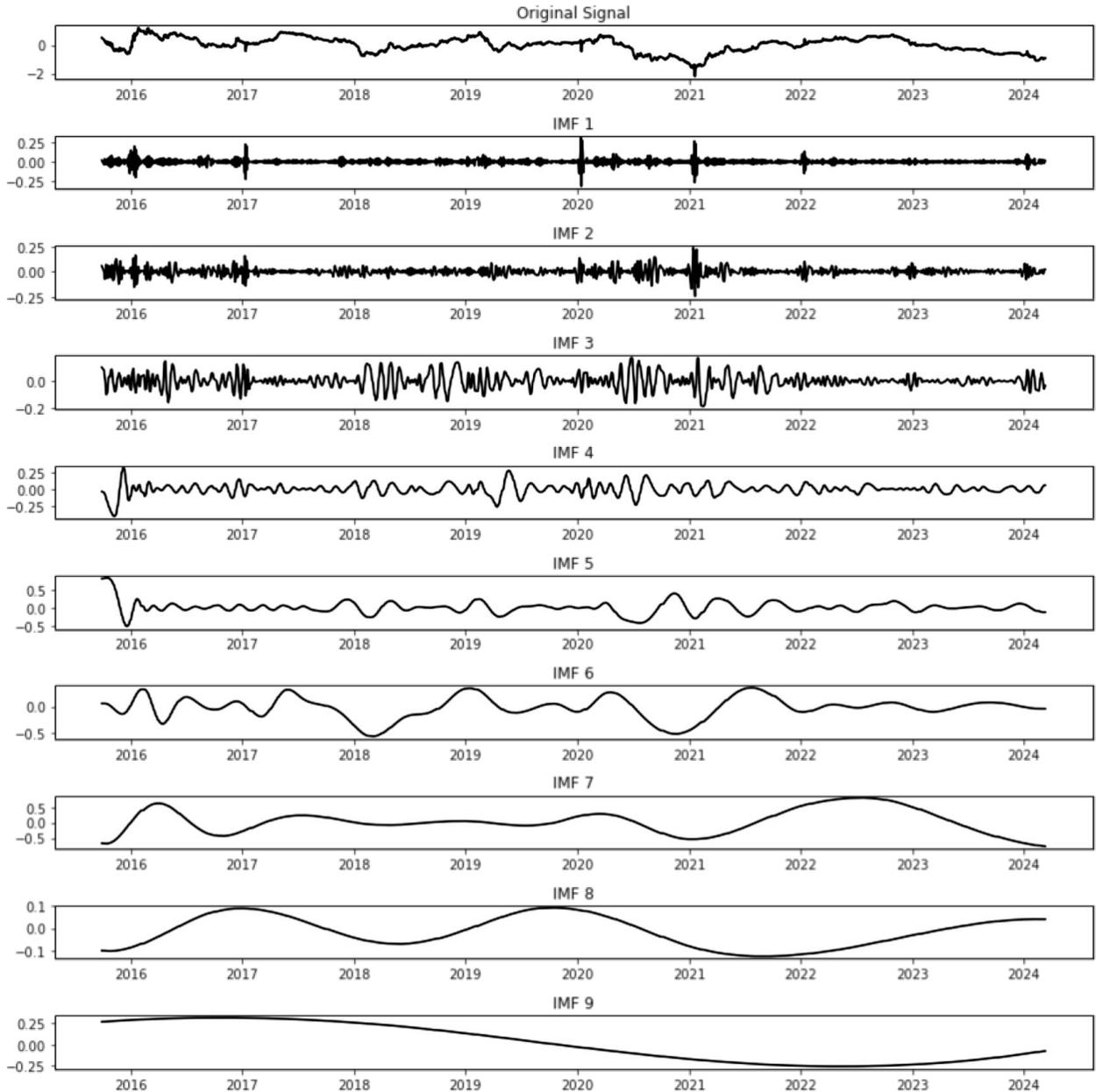

Figure 4: EMD intrinsic mode functions

### 2.3.3. Moving Average

The moving average method calculates a 22-day average as low-frequency sentiment and the daily residual as high-frequency sentiment. A significant drawback of the moving average method is its subjectivity in selecting the duration for low-frequency sentiment. Different investors may choose monthly 20-day, quarterly 60-day, or even annual moving averages. However, its advantage lies in its simplicity of operation.

## 2.4. Empirical Study

### 2.4.1. parameter form

Under these parameters, we chose a first-order lag for autoregression, and the regression results are shown in the table. We can observe that as the maturity of the options increases, the impact of long-term sentiment on the implied volatility surface parameters becomes more significant, while high-frequency sentiment only has a significant effect on options expiring within one month.

Regarding curvature, the absolute impact of high-frequency sentiment is about 5 to 10 times greater than that of low-frequency sentiment, thus playing a dominant role. For options with near-term expirations, higher short-term high-frequency market sentiment corresponds to greater curvature and a more pronounced smile. This indicates that if daily sentiment significantly deviates from the larger cycle sentiment of the period, its impact on implied volatility is positively correlated, and this correlation will be reflected in the prices of options nearing expiration. However, for options with longer expirations, a decrease in curvature is observed. The curvature correlation coefficients for options with maturities of 6 months and 12 months are -1.08 and -1.980, respectively. The influence of low-frequency sentiment is generally negative, indicating that the implied volatility smile encapsulates tail information about the price fluctuations of the underlying asset. This suggests that such information is often overestimated during periods of high short-term market sentiment, but long-term investors believe this overestimation will revert to the mean. Therefore, the pricing of options with longer expirations is negatively correlated with high-frequency market sentiment.

| $\tau$ | 1 | | 3 | | 6 | | 12 | |
|---|---|---|---|---|---|---|---|---|
| | skew | cur | skew | cur | skew | cur | skew | cur |
| $HFS_{t-1}$ | 0.194** | 0.0502 | 0.161 | 0.0155 | -0.00432 | -0.108* | -0.00232 | -1.980*** |
| | (0.095) | (0.052) | (0.101) | (0.041) | (0.145) | (0.065) | (0.220) | (0.595) |
| $LFS_{t-1}$ | -0.0183 | -0.0121 | -0.0283* | -0.0163*** | 0.0459** | -0.0297*** | 0.0961*** | -0.176** |
| | (0.014) | (0.008) | (0.015) | (0.006) | (0.021) | (0.010) | (0.032) | (0.087) |
| Constant | 0.144*** | 0.0959*** | 0.0894*** | 0.0718*** | 0.0211 | 0.0370*** | -0.0874* | -0.214* |
| | (0.020) | (0.011) | (0.021) | (0.008) | (0.030) | (0.014) | (0.046) | (0.124) |

Table 1: Parameters (skew and curvature) with different $\tau$

Regarding skewness, for the "smirk" of the implied volatility surface, options with near-term expirations are primarily influenced by high-frequency sentiment, while options with longer expirations are mainly affected by long-term sentiment. In both cases, the dominant influences are positively correlated. Higher long-term market sentiment leads to a more skewed implied volatility smirk, representing that different investors, during periods of high sentiment, are more optimistic about the downside tail risk.

For different at-the-money states, we consider the slope of this curve. It can be observed that the slope of the cross-sectional curve for near-the-money options is significantly related to high-frequency sentiment coefficients, but not to low-frequency sentiment coefficients. Conversely, the slope of the cross-sectional curve for deep out-of-the-money options is significantly related to low-frequency sentiment coefficients, but not to high-frequency sentiment coefficients, which validates our hypothesis. However, it is noteworthy that the correlation between sentiment and deep in-the-money put options with an at-the-money state of 60.0% is poor, possibly due to low liquidity.

| K | 1300 | 1100 | 1025 | 1000 | 0975 | 0900 | 0600 |
|---|---|---|---|---|---|---|---|
| $HFS_{t-1}$ | -0.00248 (0.005) | -0.0024 (0.003) | 0.00660** (0.003) | -0.0115*** (0.004) | 0.00208 (0.002) | 0.00496* (0.002) | -0.00885 (0.008) |
| $LFS_{t-1}$ | -0.0917*** (0.035) | -0.0445** (0.019) | -0.0226 (0.022) | -0.0467* (0.026) | -0.0253* (0.015) | 0.025* (0.016) | 0.0667 (0.052) |
| Constant | 0.0285*** (0.007) | 0.0142*** (0.004) | -0.00121 (0.005) | -0.00921* (0.005) | -0.0108*** (0.003) | 0.00526 (0.003) | 0.0128 (0.011) |

Table 2: Slope with different K

*2.4.2. non-parameter form*

For the non-parametric form of autoregression, we selected only 12 sets of points for the VAR model to mitigate the curse of dimensionality. These points correspond to options with strike prices of 1300, 975, and 600. We chose the option with a strike price of 975 instead of 1000 because, due to transaction costs and taxes, the behavior of call options with a strike price of 975 is closer to at-the-money options without transaction costs. After conducting an AIC test, we determined that the optimal lag order is 4. We also performed cointegration tests and Granger causality tests. The impulse response graphs are as follows:

The first row represents the response of implied volatility to shocks from high-frequency sentiment (HFS), while the second row represents the response of implied volatility to shocks from low-frequency sentiment (LFS).In the impulse response graphs mentioned above, under the impact of the high-frequency sentiment index, when K=1.3S, the effect of the high-frequency sentiment index on DOTM options dissipates after approximately 8 periods, while the high-frequency sentiment index has a negative impact on DOTM options. For near-the-money options, the effect tends to disappear around the 6th period. When K=0.6S, the impact of high-frequency sentiment on options expiring in one month disappears around the 10th period, while the impact on other DOTM options dissipates around the 6th period. From this analysis, we can conclude that the impact cycle of high-frequency sentiment on near-the-money options is relatively short, while its impact cycle on deep out-of-the-money options is longer.

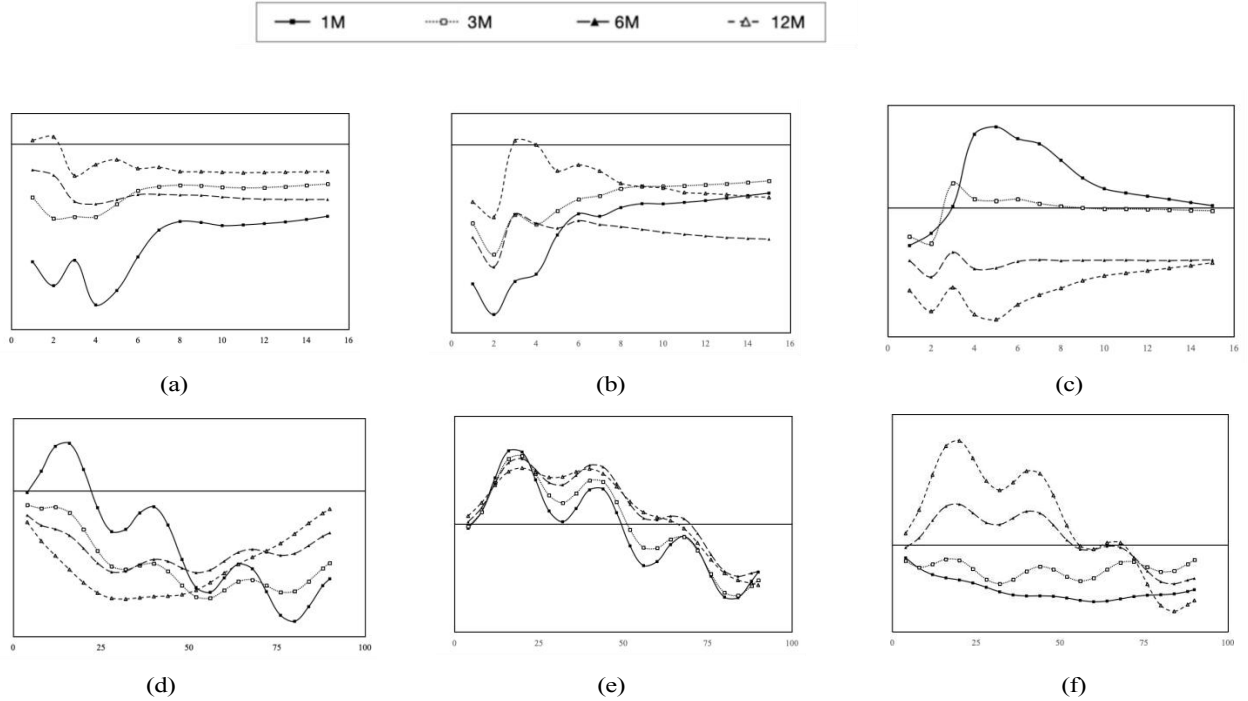

Figure 5: Impulse response graphs

For low-frequency sentiment, the research results show that the impulse effect graph does not exhibit significance. A possible explanation is that it might reflect the complexity and uncertainty of the market. The behavior of financial markets is often influenced by various factors, including economic fundamentals, political events, technical indicators, and more. Therefore, even if the sentiment index has an impact on the market in the short term, the influence of other factors might obscure or offset this impact, making the impulse effect of low-frequency sentiment statistically insignificant. Additionally, the more noticeable periodic fluctuations are due to options facing expiration every 20 trading days.

## 2.5. Robust Test

We changed the time window and used implied volatility data and market sentiment data from 2016 to 2021 to conduct vector auto-regression. This is because serious stock market crashes occurred in both 2015 and 2021. We compared this with the regression data from the previous period of 2015 to 2023 and still reached the same conclusions. Additionally, we compared the regression coefficient matrices of the two periods and found no significant changes in the significance of the independent variables.

Considering that the calculation of implied volatility is related to $\tau$ and $K$, and potentially also to the underlying price $S$ and the estimated risk-free rate $r$, we can follow Chen and Zhao (2017) by treating NBP as an exogenous variable. By controlling for variables, we can include $S$ and $r$ as exogenous variables in the VAR model. Therefore, we propose the following model:

$$Y_t = \sum_k \Phi_k Y_{t-k} + \gamma_1 S_t + \gamma_2 r_t + \varepsilon_t$$

where $Y_t$ still has the same meaning as in equation 6.

| $\tau$ | 1 | | 3 | | 6 | | 12 | |
|---|---|---|---|---|---|---|---|---|
| | (a) | (b) | (a) | (b) | (a) | (b) | (a) | (b) |
| | K=1300 | | | | | | | |
| $HFS_{t-1}$ | -3.81** | -2.47** | -1.72** | -1.354** | -0.840 | -0.899 | 0.116 | -0.0702 |
| | (-2.86) | (-2.93) | (-3.13) | (-2.67) | (-1.53) | (-1.74) | (0.12) | (-0.07) |
| $LFS_{t-1}$ | -12.72 | -66.18 | -44.55 | -69.44 | -114.0* | -138.6** | -153.4 | -104.9 |
| | (-0.10) | (-0.88) | (-0.86) | (-1.49) | (-2.26) | (-2.96) | (-1.74) | (-1.15) |
| | K=975 | | | | | | | |
| $HFS_{t-1}$ | -2.22*** | -1.773* | -1.25** | -0.837 | -1.48*** | -1.470** | -0.914 | -0.927 |
| | (-3.41) | (-2.53) | (-2.60) | (-1.55) | (-3.56) | (-3.06) | (-1.80) | (-1.56) |
| $LFS_{t-1}$ | -31.60 | -40.85 | -13.13 | -8.009 | 7.743 | 1.222 | 29.52 | 33.35 |
| | (-0.53) | (-0.65) | (-0.29) | (-0.16) | (0.20) | (0.03) | (0.63) | (0.61) |
| | K=600 | | | | | | | |
| $HFS_{t-1}$ | -1.468 | 1.139 | -1.113 | -0.338 | -2.05** | -1.979*** | -3.21** | -3.556*** |
| | (-0.67) | (1.19) | (-1.17) | (-0.56) | (-3.20) | (-3.99) | (-3.22) | (-3.80) |
| $LFS_{t-1}$ | -85.58 | -38.91 | -111.4 | -37.16 | -48.82 | -26.87 | 43.89 | 51.24 |
| | (-0.43) | (-0.46) | (-1.25) | (-0.67) | (-0.83) | (-0.60) | (0.48) | (0.60) |

Table 3: Robust test: different time period

Here, we present partial results of the coefficient matrix for the first-order lag term of the non-parametric autoregression. We can observe that for options expiring within one month, the impact of high-frequency sentiment (HFS) is significant, while only deep out-of- the-money (DOTM) options are influenced by long-term sentiment (LFS). For options with longer expiration dates, the significance of the impact from LFS gradually increases, while the significance of the impact from HFS gradually decreases. Only out-of-the-money put options show a significant correlation with HFS, consistent with the results discussed in 2.4.

## 3. Prediction of IVS

### 3.1. Machine Learning to Get Daliy Sentiment

In a news forum where investors actively participate, the daily sentiment should reflect the sentiment of the day. Loughran and Mcdonald (2011) utilized a dictionary-based approach to conduct sentiment analysis on financial texts, observing daily sentiment:

$$\Delta IS_{\text{LM}} = \frac{N_{\text{pos}} - N_{\text{neg}}}{N_{\text{total words}}}$$

In addition to the dictionary method, we trained a deep learning model to identify daily sentiment in the forum based on LSTM models and monthly BW indices. One advantage of LSTM is its ability to mimic human memory properties; investors' memory in forums is not limited to the text of the day but is also influenced by texts from previous days. Financial data typically exhibit long memory characteristics, making the LSTM method advantageous for processing financial data compared to other machine learning methods Medvedev and Wang (2022). Considering the limited nature of our training data, we employed two compar- ative methods. Firstly, we conducted data augmentation (DA) by randomly sampling text data within a month, shuffling the order. With a sufficiently large sample size, it ensures representation of the month's sentiment. Subsequently, we tokenized and encoded the top 6000 words by frequency, inputting them into the LSTM for feature learning. After train- ing, the model can be used to obtain daily sentiment from daily text inputs. Secondly, we utilized transfer learning through BERT-BigBird. By pre-training word embeddings with BERT-BigBird, followed by linear fully connected neural layers, we obtained a deep learning model with strong capabilities.

| $\tau$ | k | 1300 | 1100 | 1025 | 1000 | 975 | 900 | 600 |
|---|---|---|---|---|---|---|---|---|
| 1 | $HFS_{t-1}$ | -3.047*** | -1.781*** | -1.336*** | -1.027* | -1.734*** | -1.029** | -0.326 |
| | | (1.126) | (0.486) | (0.514) | (0.606) | (0.543) | (0.517) | (1.911) |
| | $LFS_{t-1}$ | 0.00498 | -0.0854 | -0.00674 | -0.0247 | -0.042 | 0.0864 | -0.595* |
| | | (0.201) | (0.087) | (0.092) | (0.108) | (0.097) | (0.092) | (0.341) |
| | r | -0.589*** | -0.0975 | -0.0639 | -0.0705 | -0.193** | -0.228*** | -1.299*** |
| | | (0.170) | (0.073) | (0.078) | (0.091) | (0.082) | (0.078) | (0.288) |
| | S | 0.273 | 0.282*** | 0.202* | 0.188 | -0.246** | 0.121 | -0.288 |
| | | (0.253) | (0.109) | (0.115) | (0.136) | (0.122) | (0.116) | (0.429) |
| | Constant | 2.442*** | 0.755* | 0.243 | 0.48 | 1.409*** | 1.753*** | 8.577*** |
| | | (0.910) | (0.393) | (0.415) | (0.490) | (0.439) | (0.418) | (1.546) |
| 3 | $HFS_{t-1}$ | -1.526*** | -0.759** | -0.521 | -0.693 | -1.324*** | -1.055*** | -0.675 |
| | | (0.468) | (0.327) | (0.384) | (0.671) | (0.407) | (0.374) | (0.816) |
| | $LFS_{t-1}$ | -0.245*** | -0.0861 | 0.00434 | -0.0562 | -0.109 | 0.0142 | -0.143 |
| | | (0.086) | (0.060) | (0.071) | (0.123) | (0.075) | (0.069) | (0.150) |
| | r | -0.265*** | -0.0728 | -0.128** | -0.0355 | -0.098 | -0.0920* | -0.389*** |
| | | (0.069) | (0.049) | (0.057) | (0.100) | (0.060) | (0.056) | (0.121) |
| | S | -0.0283 | 0.117 | 0.119 | 0.153 | -0.177* | 0.0382 | 0.00665 |
| | | (0.107) | (0.075) | (0.088) | (0.154) | (0.093) | (0.086) | (0.187) |
| | Constant | 1.447*** | 0.453* | 0.207 | 0.476 | 1.019*** | 0.674** | 2.276*** |
| | | (0.387) | (0.270) | (0.317) | (0.555) | (0.336) | (0.310) | (0.675) |
| 6 | $HFS_{t-1}$ | -0.615 | -0.454 | -0.492 | -0.268 | -1.298*** | -1.629*** | -2.066*** |
| | | (0.451) | (0.340) | (0.416) | (0.761) | (0.346) | (0.318) | (0.538) |
| | $LFS_{t-1}$ | -0.398*** | -0.110* | -0.0429 | 0.164 | -0.0974 | -0.0259 | -0.0703 |
| | | (0.080) | (0.061) | (0.074) | (0.136) | (0.062) | (0.057) | (0.096) |
| | r | 0.0313 | -0.0292 | -0.0765 | -0.014 | 0.00745 | -0.0221 | -0.107 |
| | | (0.063) | (0.047) | (0.058) | (0.105) | (0.048) | (0.044) | (0.075) |
| | S | -0.0204 | 0.103 | 0.0834 | 0.146 | -0.141* | -0.0122 | -0.058 |
| | | (0.100) | (0.075) | (0.092) | (0.169) | (0.077) | (0.071) | (0.119) |
| | Constant | 0.794** | 0.302 | 0.365 | 0.491 | 0.608** | 0.296 | 1.219*** |
| | | (0.354) | (0.267) | (0.326) | (0.598) | (0.272) | (0.249) | (0.423) |
| 12 | $HFS_{t-1}$ | -0.0966 | -0.21 | -0.212 | 0.648 | -0.787* | -2.026*** | -3.267*** |
| | | (0.816) | (0.638) | (0.746) | (1.088) | (0.440) | (0.450) | (0.854) |
| | $LFS_{t-1}$ | -0.311** | -0.095 | 0.0716 | 0.528*** | -0.074 | -0.165** | -0.119 |
| | | (0.142) | (0.111) | (0.130) | (0.189) | (0.077) | (0.078) | (0.148) |
| | r | -0.051 | -0.0967 | -0.155 | -0.0617 | 0.0625 | -0.0648 | -0.217* |
| | | (0.114) | (0.089) | (0.104) | (0.152) | (0.062) | (0.063) | (0.120) |
| | S | -0.151 | 0.289* | 0.370** | 0.353 | -0.0213 | -0.174* | -0.850*** |
| | | (0.189) | (0.148) | (0.172) | (0.252) | (0.102) | (0.104) | (0.198) |
| | Constant | 2.058*** | 0.678 | 1.116* | 0.968 | 0.391 | 0.692* | 3.194*** |
| | | (0.673) | (0.526) | (0.615) | (0.897) | (0.363) | (0.371) | (0.704) |

Table 4: Controlled variables parameters

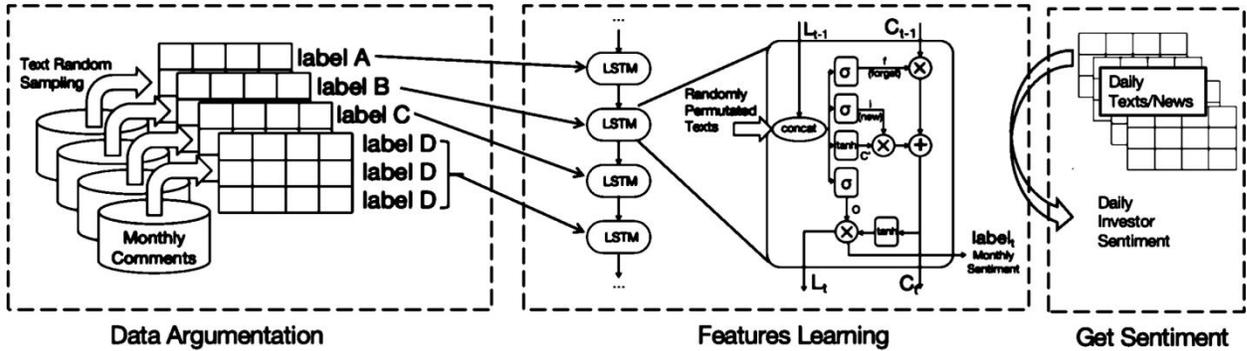

Figure 6: LSTM Method

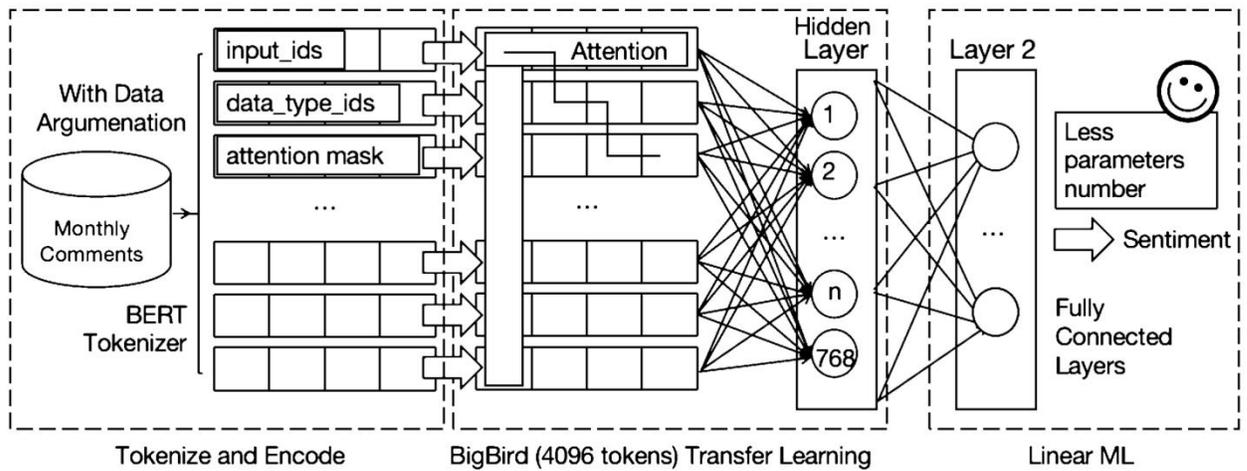

Figure 7: BigBird-Linear Method

Furthermore, we examined the impact of different sentiment acquisition methods on regression coefficients, including dictionary-based methods, principal component analysis (PCA), long short-term memory networks (LSTM), and BERT-BigBird transfer learning. High-frequency and low-frequency sentiment indices obtained through these methods were incorporated into the model. We found that for long-term deep out-of-the-money options, low-frequency sentiment had stronger explanatory power, whereas for short-term near-term options, high-frequency sentiment had stronger explanatory power. As the expiration date approached, the explanatory power of high-frequency sentiment gradually decreased while that of low-frequency sentiment gradually increased, especially for deep out-of-the-money options expiring in March, where both high-frequency and low-frequency sentiments exhib- ited strong explanatory power.

*3.2. Predict Accuracy*

In the table, we summarize the Mean Squared Prediction Errors (MPSE) of applying various sentiment extraction methods combined with VAR models for forecasting implied volatility surfaces. These methods include long-term high-frequency sentiments extracted from dictionaries, PCA method, LSTM training, and BBL transfer learning, and their pre- dictive effects on implied volatility for options with different expiration dates.

Overall, except for the dictionary-based sentiment which performs worse than the autore- gressive model without any sentiment, the other methods show improvements in predictive performance. Among them, the BERT-BigBird model performs best for options with 1- month, 3-month, and 12-month expirations. The superiority of the PCA method over the LSTM method might be due to overfitting during LSTM training. Although we augmented the data during LSTM model setup, the parameter size of LSTM remained at around 140,000, which is not easily avoidable relative to our dataset containing 610,000 text samples and 116 monthly labels. In contrast, BERT-BigBird transfer learning, due to the restriction on training word vectors, did not exhibit significant overfitting and showed better predictive performance compared to LSTM.

|       | None   | Dictionary | PCA    | LSTM   | BERT-BigBird |
|-------|--------|------------|--------|--------|--------------|
| 1M    | 21.13% | 21.05%     | 21.59% | 21.17% | 20.83%       |
| 3M    | 16.65% | 16.69%     | 16.34% | 16.62% | 16.19%       |
| 12M   | 28.83% | 29.08%     | 28.30% | 28.79% | 28.09%       |
| Tatal | 22.21% | 22.27%     | 22.08% | 22.19% | 21.70%       |

Table 5: Prediction accuracy

Compared to Medvedev and Wang (2022)who predicted with US S&P500 options data, with a VAR model prediction accuracy of 13.72%, and the Convolutional Long Short-Term Memory (ConvLSTM) reaching an optimal of 8.26%, our optimal prediction accuracy using Shanghai Stock Exchange 50ETF data is 16.19%. The data difference may partially be attributed to the characteristics of the Chinese options market and partly to the fact that our deep learning component only focused on sentiment analysis of news to maintain model interpretability. Our prediction errors are in the same order of magnitude as theirs.

The figure shows the predictive performance of various sentiment analysis methods on the implied volatility surface. To illustrate the "smirk" shape of the volatility surface in two dimensions, we use the value state as the horizontal axis. It can be observed that the predictions of the dictionary method are more concentrated, while the LSTM and BERT models have broader coverage.

This may be due to the Chinese-specific "context" and non-standardized text. For ex- ample, in February 2024, a stock investor posted on a forum,"Really good, saved money for traveling during the New Year ".The essence of this statement is closer to negative sentiment, but words like "good" and "saved" are identified by the financial dictionary as positive, re- sulting in a sentiment score close to zero for that day. Similarly, in August 2023, a stock investor posted on a forum, "Hold on! GJD is about to take action [emoji] " If analyzed according to the dictionary, neither "GJD" nor the emoji are included in the LM Chinese dictionary. While "Hold on" is recognized as a positive word, "take action" has the same characters as "sold out" is mistakenly classified as negative, resulting in a sentiment score of zero for the statement. Such errors lead to predictions that are concentrated towards the middle range.

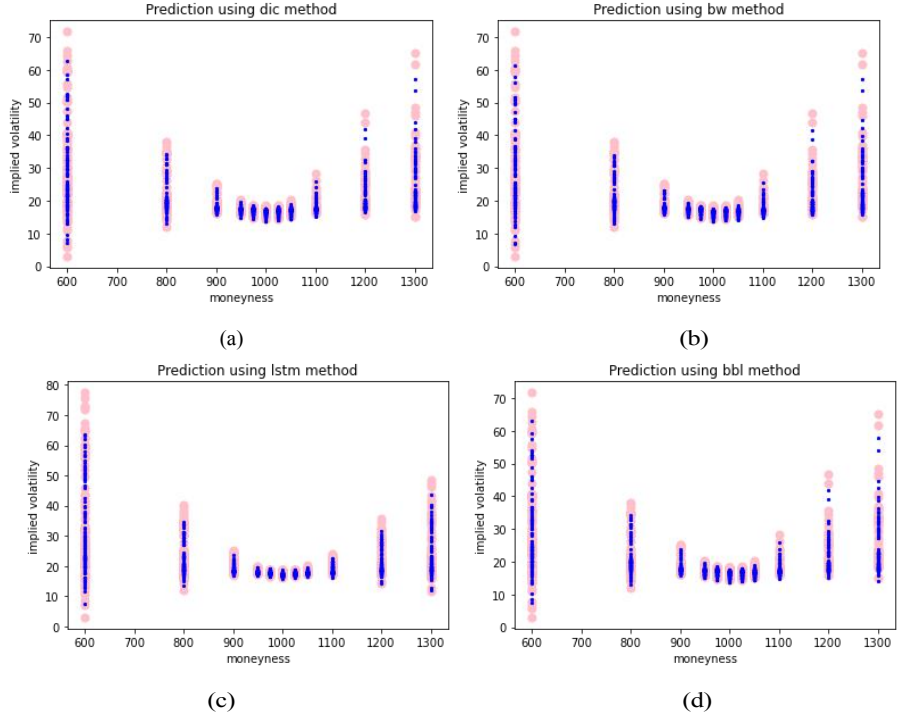

Figure 8: IVS: Real value and predicted value

For the above issue, the LSTM method and BERT transfer learning model are able to effectively avoid it. This is because they recognize all textual information and, combined with the provided labels for market sentiment, they can learn the partial features of out-of- dictionary words in the financial domain. For words unrelated to the domain, they can also promptly iterate through forgetting dropout functions, ensuring better performance when predicting extreme values.

## 4. Conclusion and Further Discussion

Through the integration of natural language processing (NLP) techniques and deep learn- ing models, we have delved into the relationship between market sentiment and implied volatility surface, proposing a novel prediction model. Leveraging a dataset of 630,000 tex- tual records from the East Money Stock Forum spanning from 2014 to 2023, we constructed daily market sentiment indicators using deep learning methods such as BERT and LSTM. Our research reveals that short-term sentiment correlates more strongly with at-the-money (ATM) implied volatility of options, while long-term sentiment exhibits a more significant correlation with deep out-of-the-money (DOTM) implied volatility.

In our empirical study, we employed a Vector Autoregression (VAR) model incorporating sentiment indicators into the prediction framework, significantly improving the accuracy of predicting next-day implied volatility surfaces. Particularly noteworthy is the utilization of the BERT-BigBird model for sentiment analysis, which achieved optimal Mean Percentage Absolute Error (MPAE) for implied volatility predictions across various option maturities of 1, 3, and 12 months, with percentages of 20.83%, 16.19%, and 28.09%, respectively. This indicates the model's high accuracy in forecasting implied volatility for options with different tenors.

Regarding robustness testing, we shortened the time window to 2015-2021 and compared it with data from 2015-2023, validating the stability of the model's prediction results. Con- sistency in regression coefficients across different time periods further confirms the reliability and robustness of the model.

However, our empirical findings indicate that for options expiring in January, with an ATM state of 60.0%, and surrounding put options, both high-frequency and low-frequency sentiments generally do not have significant explanatory power. One conjecture is that this may be due to their weaker liquidity, with fewer investors participating in trading these options, thus failing to generate significant behavioral finance effects. Further exploration is required to delve into the underlying reasons.